\documentclass[10pt,twocolumn,letterpaper]{article}

\usepackage{cvpr}
\usepackage{times}
\usepackage{epsfig}
\usepackage{graphicx}
\usepackage{amsmath}
\usepackage{amssymb}
\usepackage{bm}
\usepackage{caption}
\usepackage{color}
\usepackage{enumitem,multirow}
\usepackage{relsize,cite}
\DeclareMathAlphabet{\mathpzc}{T1}{pzc}{m}
{it}

\newcommand{\R}{{\mathbb{R}}}

\newcommand{\norm}[1]{\left\lVert#1\right\rVert}

\usepackage[pagebackref=true,breaklinks=true,letterpaper=true,colorlinks,bookmarks=false]{hyperref}

\cvprfinalcopy 

\def\cvprPaperID{3760} 

\ifcvprfinal\fi

\title{Depth from a polarisation + RGB stereo pair}

\author{Dizhong Zhu and William A. P. Smith\\
University of York, York, UK\\
{\tt\small \{dz761,william.smith\}@york.ac.uk}
}

\begin{document}
\twocolumn[{%
\renewcommand\twocolumn[1][]{#1}%

\maketitle
\thispagestyle{empty}

\begin{center}
    \centering
    \includegraphics[width=\textwidth,clip=true]{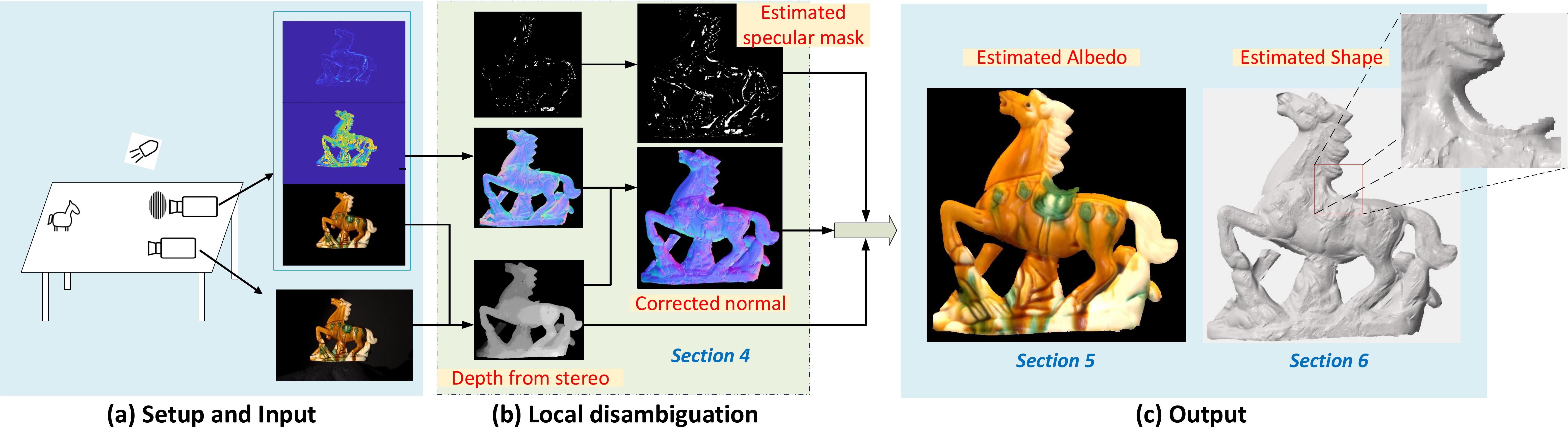}
\vspace{-0.6cm}
    \captionof{figure}{Overview: From a stereo pair of one polarisation image and one RGB image (a) we merge stereo depth with polarisation normals using a higher order graphical model (b) before estimating an albedo map and the final geometry (c).}\label{fig:teaser}
\end{center}%
}]

\begin{abstract}
In this paper, we propose a hybrid depth imaging system in which a polarisation camera is augmented by a second image from a standard digital camera. For this modest increase in equipment complexity over conventional shape-from-polarisation, we obtain a number of benefits that enable us to overcome longstanding problems with the polarisation shape cue. The stereo cue provides a depth map which, although coarse, is metrically accurate. This is used as a guide surface for disambiguation of the polarisation surface normal estimates using a higher order graphical model. In turn, these are used to estimate diffuse albedo. By extending a previous shape-from-polarisation method to the perspective case, we show how to compute dense, detailed maps of absolute depth, while retaining a linear formulation. We show that our hybrid method is able to recover dense 3D geometry that is superior to state-of-the-art shape-from-polarisation or two view stereo alone.
\end{abstract}

\section{Introduction}

Surface reflection changes the polarisation state of light. By measuring the polarisation state of reflected light, we are able to infer information about the material properties and geometry of the surface. Polarisation is a particularly attractive shape estimation cue because it is dense (surface orientation information is available at every pixel), can be applied to smooth, featureless, glossy surfaces (on which multiview methods would fail to find correspondences) and it can be captured in a single shot (using a polarisation camera). For this reason, the shape-from-polarisation cue has recently been rediscovered and significant progress has been made in the past three years \cite{tozza2017linear,smith2016linear,kadambi2015polarized,kadambi2017depth,cui2017polarimetric,ngo2015shape,berger2017depth,atkinson2017polarisation,yu2017shape,mecca2017differential}.

Recent work has posed shape-from-polarisation in terms of direct estimation of orthographic surface height \cite{tozza2017linear,smith2016linear,smith2018height}. This is attractive because it halves the degrees of freedom (one height value per pixel rather than two values to represent surface orientation) and avoids the two step process of surface orientation estimation followed by surface integration to obtain a height map. However, polarisation cues do not provide any direct constraints on metric depth, only on local surface orientation. Hence, the surfaces recovered by these methods are globally inaccurate and subject to low frequency distortion. Moreover, the orthographic assumption is practically limiting.

For this reason, in this paper we consider a hybrid setup in which a single polarisation image is augmented by a second image from a standard RGB camera. This provides us with a conventional stereo cue from which we can compute coarse but metrically accurate depth estimates. This serves a number of purposes. First, this provides coarse guide normals that can be used for initial disambiguation of the polarisation cue. Second, it is used to regularise the final reconstruction, resolving scale ambiguity and reducing low frequency bias. We make a number of novel contributions:
\begin{enumerate}[leftmargin=1.2em]
    \setlength{\itemsep}{0pt}
    \setlength{\parskip}{0pt}
    \item Use a higher order graphical model to capture integrability constraints during disambiguation
    \item Show how to automatically label pixels as diffuse or specular dominant via our graphical model
    \item Show how to incorporate gradient-consistency constraints into albedo estimation
    \item Extend the linear formulation of Smith \etal \cite{smith2016linear} to the perspective case, retaining linearity and also including the stereo depth map as a guide surface
\end{enumerate}
Our approach has a number of practical advantages over recent state-of-the-art. Unlike Smith \etal \cite{smith2016linear} we do not assume uniform albedo. Unlike Kadambi \etal \cite{kadambi2015polarized,kadambi2017depth}, we do not use a depth (kinect) camera and so our capture environment is not restricted. We compare to these and other relevant state-of-the-art methods and obtain better reconstructions. Compared to \cite{chen2018polarimetric,berger2017depth,yang2018polarimetric,cui2017polarimetric}, we only require a single polarisation image. 

\subsection{Related work}

\noindent \textbf{Shape-from-polarisation.}  
Both Miyazaki \etal \cite{miyazaki2003polarization} and Atkinson and Hancock \cite{atkinson2006recovery} used a diffuse polarisation model to estimate surface normals from the phase angle and degree of polarisation. They use a local, greeedy method that propagates from the object boundary assuming global convexity. This is very sensitive to noise, limits applicability to objects with a visible occluding boundary and does not consider integrability. Morel \etal \cite{morel2005polarization} took a similar approach but used a specular polarisation model suitable for metallic surfaces. Huynh \etal \cite{huynh2010shape} also assumed convexity to disambiguate the polarisation normals.

\noindent \textbf{Polarisation and X.} A variety of work seeks to augment polarisation with an additional shape-from-X cue. Huynh \etal \cite{huynh2013shape} extended their earlier work to use multispectral measurements to estimate both shape and refractive index. Drbohlav and Sara \cite{drbohlav2001unambiguous} showed how the Bas-relief ambiguity \cite{belhumeur1999bas} in uncalibrated photometric stereo could be resolved using polarisation. However, this approach requires a polarised light source. Coarse geometry obtained by multi-view space carving \cite{miyazaki2012polarization,miyazaki2017surface} has been used to resolve polarisation ambiguities. Kadambi \etal \cite{kadambi2015polarized,kadambi2017depth} combine a single polarisation image with a depth map obtained by an RGBD camera. The depth map is used to disambiguate the normals and provide a base surface for integration. Our approach uses a simpler setup in that it does not require a depth camera.
Mahmoud \etal \cite{mahmoud2012direct} and Smith \etal \cite{smith2016linear} augment polarisation with a shape-from-shading cue. The later shows how to solve directly for surface height (i.e. relative depth) by solving a large, sparse linear system of equations. However, they assume constant albedo and orthographic projection - all assumptions that we avoid. Follow-up work showed how to estimate albedo independently \cite{smith2018height}. Yu \etal \cite{yu2017shape} take a similar approach but avoid linearising the objective function, instead directly minimising the true nonlinear objective. This allows the use of reflectance and polarisation models of arbitrary complexity. Ngo \etal \cite{ngo2015shape} derived constraints that allowed surface normals, light directions and refractive index to be estimated from polarisation images under varying lighting. However, this approach requires at least 4 light directions. Atkinson \cite{atkinson2017polarisation} combine calibrated two source photometric stereo with polarisation phase and resolve ambiguities via a region growing process. Tozza \etal \cite{tozza2017linear} generalised \cite{smith2016linear} to consider two source photo-polarimetric shape estimation. Subsequently, Mecca \etal \cite{mecca2017differential} also proposed a differential formulation with a well-posed solution for two light sources.

\noindent \textbf{Multiview Polarisation.} Some of the earliest work on polarisation vision used a stereo pair of polarisation measurements to determine the orientation of a plane \cite{wolff1990surface}. Rahmann and Canterakis \cite{rahmann2001reconstruction} combined a specular polarisation model with stereo cues. Similarly, Atkinson and Hancock \cite{atkinson2007shape} used polarisation normals to segment an object into patches, simplifying stereo matching. Note however that this method is restricted to the case of an object rotating on a turntable with known angle. Stereo polarisation cues have also been used for transparent surface modelling \cite{miyazaki2004transparent}. Berger \etal \cite{berger2017depth} used polarisation stereo for depth estimation of specular scenes. Cui \etal \cite{cui2017polarimetric} incorporate a polarisation phase angle cue into multiview stereo enabling recovery of surface shape in featureless regions. Chen \etal \cite{chen2018polarimetric} provide a theoretical treament of constraints arising from three view polarisation. Yang \etal \cite{yang2018polarimetric} propose a variant of monocular SLAM using polarisation video. All of these methods require multiple polarisation images whereas our proposed approach uses only a single polarisation image augmented by a standard RGB image from a second view.

\section{Problem formulation}

In this section we list our assumptions and introduce notations, the perspective surface depth representation and basic polarisation theory.

\subsection{Assumptions}

Our method makes the following assumptions:
\begin{itemize}[leftmargin=1.2em]
    \setlength{\itemsep}{0pt}
    \setlength{\parskip}{0pt}
    \item Intrinsic parameters of both cameras known
    \item Dielectric material with known refractive index
    \item Distant point light source with known direction
    \item Diffuse reflectance follows Lambert's law
    \item Object is smooth, i.e. $C^2$-continuous (integrable)
\end{itemize}
These assumptions are all common to previous work. We draw attention to the fact that we do not assume orthographic projection, known albedo or that pixels have been labelled as diffuse or specular dominant, making our approach more general than previous work.

\subsection{Perspective depth representation}

Our setup consists of a polarisation camera and an RGB camera. We work in the coordinate system of the polarisation camera and parameterise the surface by the unknown depth function $Z(\mathbf{u})$, where $\mathbf{u}=(x,y)$ is a location in the polarisation image. The 3D coordinate at $\mathbf{u}$ is given by:
\begin{equation} \label{eq:3Dcrsdepth}
    P(\mathbf{u}) = 
    \begin{bmatrix} 
    \frac{x-x_0}{f}Z(\mathbf{u})\\
    \frac{y-y_0}{f}Z(\mathbf{u})\\
    Z(\mathbf{u})
    \end{bmatrix},
\end{equation}
where $f$ is the focal length of the polarisation camera in the $x$ and $y$ directions and $(x_0,y_0)$ is the principal point. The direction of the outward pointing surface normal is defined as the cross product of the partial derivatives with respect to $x$ and $y$ \cite{graber2015efficient}:
\begin{equation} \label{eq:perspectivenormal}
	\mathbf{n}(\mathbf{u})=\begin{bmatrix}
		-\frac{Z(\mathbf{u})\cdot Z_x(\mathbf{u})}{f_y}\\
		-\frac{Z(\mathbf{u})\cdot Z_y(\mathbf{u})}{f_x}\\
		\frac{x-x_0}{f_x}\frac{Z(\mathbf{u})\cdot Z_x(\mathbf{u})}{f_y}+\frac{y-y_0}{f_y}\frac{Z(\mathbf{u})\cdot Z_y(\mathbf{u})}{f_x}+\frac{Z(\mathbf{u})^2}{f_xf_y}
	\end{bmatrix}
\end{equation}
where $Z_x, Z_y$ denotes the partial derivative of $Z(\mathbf{u})$ w.r.t. x and y. Note that the magnitude of $\mathbf{n}(\mathbf{u})$ is arbitrary, only its direction is important. For this reason, we can cancel any common factors. In particular, we can divide through by $Z(\mathbf{u})$ to remove quadratic terms and multiply through by $f_xf_y$ to avoid numerical instability caused by division by $f_xf_y$ (which is potentially very large):
\begin{equation} \label{eq:perspectivenormal}
	\mathbf{n}(\mathbf{u})=\begin{bmatrix}
		-f_yZ_x(\mathbf{u})\\
		-f_xZ_y(\mathbf{u})\\
		(x-x_0)Z_x(\mathbf{u}) + (y-y_0)Z_y(\mathbf{u}) +Z(\mathbf{u})
	\end{bmatrix}
\end{equation}
We denote by $\mathbf{\Bar{n}}(\mathbf{u})=\mathbf{n}(\mathbf{u})/\|\mathbf{n}(\mathbf{u})\|$, the unit length surface normal.

The vector pointing towards the viewer from a point on the surface is given by:
\begin{equation}
    \mathbf{v}(\mathbf{u}) = -\begin{bmatrix} 
    \frac{x-x_0}{f_x}&
    \frac{y-y_0}{f_y}&
    1
    \end{bmatrix}^T \mathbin{/} \left\|\begin{bmatrix} 
    \frac{x-x_0}{f_x}&
    \frac{y-y_0}{f_y}&
    1
    \end{bmatrix}\right\|.
\end{equation}
Note that this is independent of surface depth.

\subsection{Polarisation theory}\label{sec:poltheory}
When unpolarised light is reflected by a surface it becomes partially polarised~\cite{wolff1997polarization}. The polarisation information can be estimated by capturing a sequence of images in which a linear polarising filter mounting on camera lens is rotated through a sequence of $P\geq 3$ different angles $\vartheta_j$, $j\in \left\{1,\ \dots,\ P\right\}$. The measured intensity at a pixel varies sinusoidally with the polariser angle, it can be written as:
\begin{equation}\label{eqn:TRS}
 i_{\vartheta_j}(\mathbf{u})=i_{\textrm{un}}(\mathbf{u}) \, \big( 1 + \rho(\mathbf{u}) \cos(2\vartheta_j-2\phi(\mathbf{u}))\big). 
\end{equation}
The polarisation image is thus obtained by decomposing the sinusoid at every pixel location  into three quantities~\cite{wolff1997polarization}: the {\it phase angle}, $\phi(\mathbf{u})$, the {\it degree of polarisation}, $\rho(\mathbf{u})$, and the {\it unpolarised intensity}, $i_{\textrm{un}}(\mathbf{u})$.  
The parameters of the sinusoid can be estimated from the captured image sequence using non-linear least squares~\cite{AH2006}, linear methods~\cite{huynh2010shape} or via a closed form solution~\cite{wolff1997polarization} for the specific case of $P=3$, $\vartheta\in\{ 0^{\circ}, 45^{\circ}, 90^{\circ}\}$.

A polarisation image provides a constraint on the surface normal direction at each pixel. 
The exact nature of the constraint depends on the polarisation model used.
In this paper we will consider diffuse polarisation, due to subsurface scattering (see \cite{AH2006} for more details), and specular polarisation due to direct reflection.

\paragraph{Degree of polarisation constraint.}
The degree of diffuse polarisation $\rho_d(\mathbf{u})$ at each point $\mathbf{u}$ can be expressed in terms of the refractive index $\eta$ and, in the perspective case, the viewing angle $\theta(\mathbf{u}) = \arccos\left[\mathbf{\Bar{n}}(\mathbf{u})\cdot\mathbf{v}(\mathbf{u})\right] \in [0, \frac{\pi}{2}]$ as follows (Cf.~\cite{AH2006}):
\small
\begin{eqnarray} \label{eq:diffuse_degree_pol}
&&\rho_d(\mathbf{u}) =  \\
&&\hspace{-0.8cm} \frac{(\eta-1/\eta)^2 \sin^2\theta(\mathbf{u})}{2\! +\! 2\eta^2 \!-\! (\eta \!+\! 1/\eta)^2 \sin^2\theta(\mathbf{u}) \! +\! 4 \cos\theta(\mathbf{u}) \sqrt{\eta^2 \!- \sin^2 \theta(\mathbf{u})}}. \nonumber
\end{eqnarray}
\normalsize
This expression can be inverted. From the measured degree of polarisation, the viewing angle $\theta(\mathbf{u})$ (and hence one degree of freedom of the surface normal) can be estimated by rewriting  \eqref{eq:diffuse_degree_pol} \cite{smith2016linear}. This relates the cosine of the viewing angle to a function, $f(\rho(\mathbf{u}),\eta)$, that depends on the measured degree of polarisation and the refractive index:
\small
\begin{align}
&\hspace{0.cm} \cos\theta(\mathbf{u}) = \mathbf{n}(\mathbf{u})\cdot\mathbf{v}(\mathbf{u}) = f(\rho(\mathbf{u}),\eta) =  \label{eqn:DOPcon} \\
&\sqrt{\frac{\eta^4 \!(1\!-\!\rho_d^2) \!+\! 2\eta^2 (2\rho_d^2 \!+\!\rho_d \!-\!1)\! +\! \rho_d^2 \!+\! 2\rho_d \!-\!4 \eta^3 \rho_d \sqrt{1\!-\!\rho_d^2} \!+\!1}
{(\rho_d+1)^2 \, (\eta^4 + 1) + 2\eta^2(3\rho_d^2 + 2\rho_d -1)}}  \nonumber
\end{align}
\normalsize
where we drop the dependency of $\rho_d$ on $({\mathbf u})$ for brevity. Similarly, the degree of polarisation of a specular reflection is given by:
\begin{equation} \label{eq:PolRefSpec}
\begin{aligned}
\rho_s(\mathbf{u})&=\frac{2\sin^2\theta(\mathbf{u}) \cos\theta(\mathbf{u})\sqrt{\eta^2-\sin^2\theta(\mathbf{u})}}
{\eta^2-\sin^2\theta(\mathbf{u})-\eta^2\sin^2\theta(\mathbf{u})+2\sin^4\theta(\mathbf{u})}.
\end{aligned}	
\end{equation}
This expression has two solutions possible solutions for $\theta(\mathbf{u})$ given a measured degree of specular polarisation.

\paragraph{Phase angle constraint}
The phase angle determines the azimuth angle of the surface normal $\alpha({\bf u})\in [0,2\pi]$ up to a $180^{\circ}$ ambiguity. For diffuse dominant reflectance this is given by:
\begin{equation}\label{eqn:phase}
 \alpha({\bf u}) = \phi({\bf u})\ \textrm{or}\ (\phi({\bf u}) + \pi),
\end{equation}
and for specular dominant reflectance by:
\begin{equation}\label{eqn:phase}
 \alpha({\bf u}) = \phi({\bf u})\pm \frac{\pi}{2}.
\end{equation}

\paragraph{Diffuse shading constraint}
Under the assumption of perfect diffuse reflectance, the unpolarised intensity for diffuse dominant pixels follows Lambert's law:
\begin{equation}
    i_{d}(\mathbf{u}) = \frac{a(\mathbf{u})\mathbf{n}(\mathbf{u})\cdot\mathbf{s}}{\|\mathbf{n}\|},\label{eqn:lamblaw}
\end{equation}
where $\mathbf{s}\in\R^3$ is the known distant point source direction and $a(\mathbf{u})\in[0,1]$ the diffuse albedo at pixel $\mathbf{u}$.

\paragraph{Diffuse/specular dominance}

We assume that total reflectance is a mixture of subsurface diffuse reflectance, $i_d$, and specular surface reflection, $i_s$ (for which we do not assume any particular reflectance model). This means that observed sinusoid is a sum of two sinusoids with a phase difference of $\pi/2$. The resulting sinusoid will be in phase with either the diffuse or specular sinusoid depending on which reflectance ``dominates''. Concretely, if $i_d\rho_d>i_s\rho_s$ then the pixel is diffuse dominant and we neglect specular reflectance, i.e.~we assume $i_{\textrm{un}}=i_d$. 

\section{Overview of method}

Our proposed method comprises the following steps:
\begin{enumerate}[leftmargin=1.2em]
    \setlength{\itemsep}{0pt}
    \setlength{\parskip}{0pt}
    \item Estimate the disparity from stereo images and reconstruct a coarse depth map by known camera matrix.
    \item Compute guide surface normals by taking the gradient of the coarse depth map.
    \item Use guide surface normal to disambiguate the polarisation normals via a higher order graphical model.
    \item Estimate diffuse albedo from disambiguated polarisation normals.
    \item Linearly estimate perspective depth from polarisation using coarse depth map as a constraint.
\end{enumerate}
Our pipeline is illustrated in Fig.~\ref{fig:teaser} and each step is described in detail in the following sections.

\section{Integrability-based disambiguation with a higher order graphical model}
\label{sec:graph}

The constraints in Section \ref{sec:poltheory} restrict the surface normal at a pixel to six possible directions. If the pixel is diffuse dominant, then the viewing angle is uniquely determined by the degree of polarisation and the azimuth angle restricted to two possibilities by the phase angle, leading to two possible normal directions. If the pixel is specular dominant, the degree of polarisation restricts the viewing angle to two possibilities, with the azimuth again also restricted to two, given four possible normal directions in total. Previous work \cite{kadambi2015polarized, smith2016linear} assumes that the labelling of pixels as specular or diffuse dominant is known in advance. We do not assume that the labels are known and propose an initial resolution of this six-way ambiguity using a higher order graphical model. The motivation for using a higher order model is that a ternary potential can measure deviation from integrability.

\begin{figure}[t]
    \centering
    \includegraphics[width=8.5cm,clip=true]{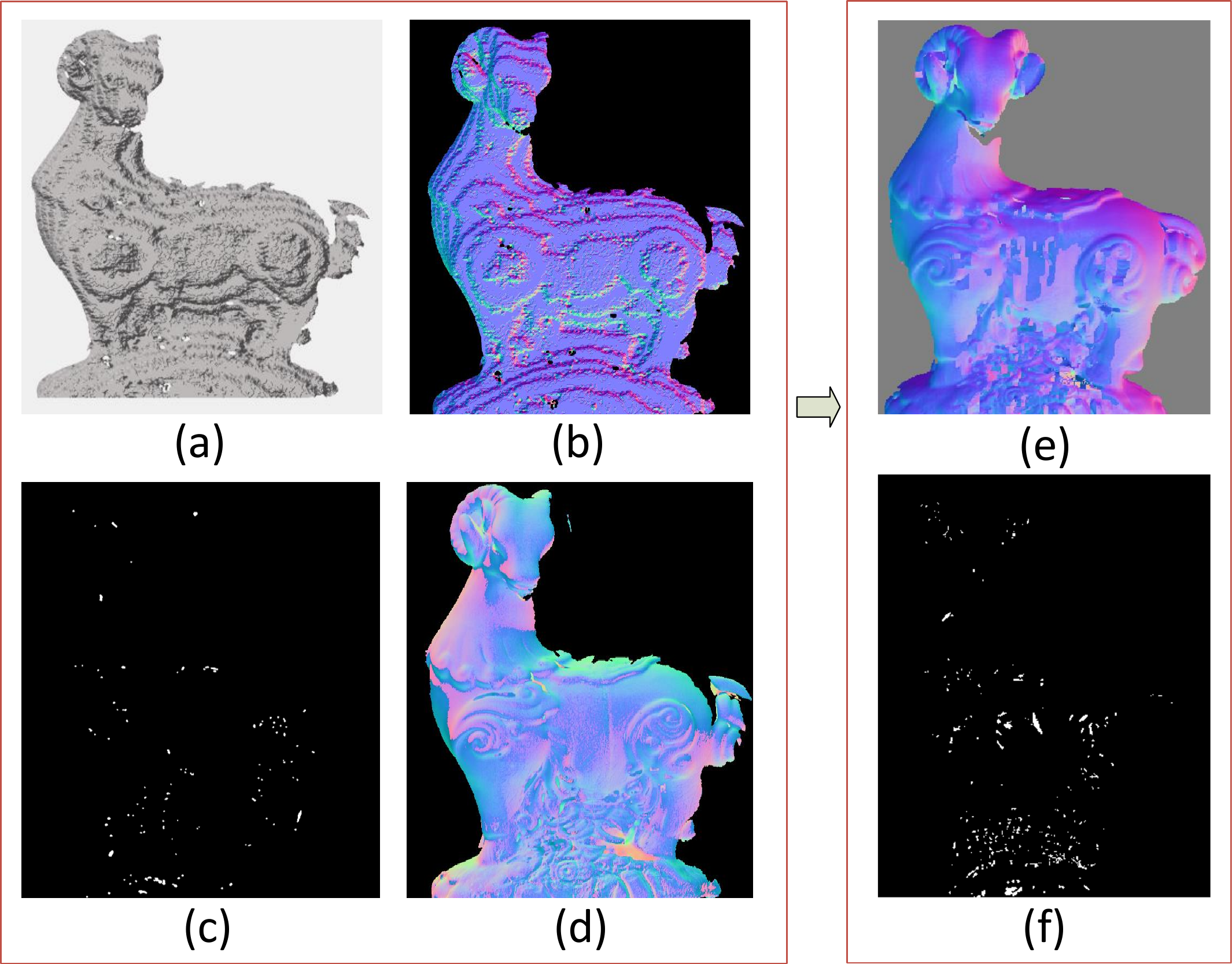}
    \caption{(a) Depth map from disparity map. (b) Guide surface normal from stereo depth map. (c) Preset specular mask. (d) One possible polarisation normal. (e) The corrected normal via our graphical model. (f) The updated specular mask via graphical model.}
    \label{fig:Normal_disambiguation}
\end{figure}

We set up an energy cost function to be mimised w.r.t.~the surface normal as follows:
\begin{equation}
\begin{aligned}
    E(\mathbf{n}(\mathbf{u}))&=\sum_{\mathbf{u}\in \mathlarger{\nu} }\Phi(\mathbf{n}(\mathbf{u}))+\sum_{\mathbf{(u,v)} \in \mathpzc{N}}\varphi(L(\mathbf{u}),L(\mathbf{v}))\\
    &+\sum_{(\mathbf{u,v,w})\in \mathpzc{T}}\Psi(\mathbf{n}(\mathbf{u}),\mathbf{n}(\mathbf{v}),\mathbf{n}(\mathbf{w}))
\end{aligned}\label{eqn:GMcost}
\end{equation}
Here $\mathlarger{\nu}$ corresponds to all foreground pixels, $\mathpzc{N}$ is the set of adjacent pixels and $\mathpzc{T}$ is the set of pixel triplets $(\mathbf{u},\mathbf{v},\mathbf{w})$ where $\mathbf{u}=(x,y)$, $\mathbf{v}=(x+1,y)$ and $\mathbf{w}=(x,y+1)$.
Before further explaining the energy terms, let us clarify two important elements that will be used in following. \textbf{1)}. The stereo setup produces a coarse depth map by computing the disparity from the camera pair. We use the semi-global matching method \cite{hirschmuller2005accurate} to compute the disparity and reconstruct a depth map with the camera matrices, as displayed in Figure \ref{fig:Normal_disambiguation}(a). Thus its surface normal can be computed by simply taking the forward difference on the coarse depth map. We denote these surface normal by $\hat{\mathbf{n}}$ where they are noisy as shown in Figure \ref{fig:Normal_disambiguation}(b). \textbf{2)}. We make a rough initial estimate of the specular/diffuse dominant pixel labelling, $L$. We simply set $L(\mathbf{u})=1$ if the measured intensity is saturated (Figure \ref{fig:Normal_disambiguation}(c)). $L$ will be subsequently updated (Figure \ref{fig:Normal_disambiguation}(f)). 
\paragraph{Unary cost}
The unary term aims to minimise the angle between $\mathbf{n}(\mathbf{u})$ and $\mathbf{\hat{n}}(\mathbf{u})$, where $\mathbf{n}(\mathbf{u})$ has up to six solutions. We denote the first two solutions from diffuse component in $\mathpzc{D}$ and the rest from specular component in $\mathpzc{S}$. We also take account the initial specular mask $L$ i.e. Where the diffuse normal will be assigned to low probability if its corresponding specular mask equal to one. The unary cost can be written as
\small{
\begin{equation}
    \begin{aligned}
    \nonumber &\Phi(\mathbf{n}(\mathbf{u}))= & \\
    &\begin{cases}
            k\cdot f(\mathbf{u}) &\text{if}\ (L(\mathbf{u})=1, \mathbf{n}(\mathbf{u})\in \mathpzc{D})\ \text{or}\ (L(\mathbf{u})=0, \mathbf{n}(\mathbf{u})\in \mathpzc{S}) \\
            f(\mathbf{u})   &\text{if}\ (L(\mathbf{u})=0, \mathbf{n}(\mathbf{u})\in \mathpzc{D})\ \text{or}\ (L(\mathbf{u})=1, \mathbf{n}(\mathbf{u})\in \mathpzc{S})
    \end{cases}
\end{aligned}
\end{equation}
}
where $f(\mathbf{u})$ depends on the cosine of the angle between $\mathbf{n}(\mathbf{u})$ and $\mathbf{\hat{n}}(\mathbf{u})$ and is defined as
\begin{equation}
    f(\mathbf{u})=\exp(-\mathbf{n}(\mathbf{u})\cdot \mathbf{\hat{n}}(\mathbf{u})).
\end{equation}
The parameter $k<1$ penalises surface normal disambiguations that are not consistent with the corresponding specular mask. We set $k=0.1$ in our experiments.

\paragraph{Pairwise cost} We encourage pairwise pixels in $\mathpzc{N}$ to have similar diffuse or specular labels and penalise where the labels changed. We define 
\begin{equation}
    \varphi(L(\mathbf{u}),L(\mathbf{v}))=|L(\mathbf{u})-L(\mathbf{v})|.
\end{equation}

\paragraph{Ternary cost} In order to encourage the disambiguated surface normals to satisfy the integrability constraint, we use a ternary cost to measure deviation from integrability. For an integrable surface, the mixed second order partial derivatives on the gradient field should be equal \cite{petrovic2001enforcing}. Specifically, $\frac{\partial p}{\partial y}=\frac{\partial q}{\partial x}$. Where $p,q$ are the partial derivatives in the $x$ and $y$ direction respectively. The surface gradient is directly linked to the surface normal by
\[
    p(\mathbf{u})=-n_x(\mathbf{u})/n_z(\mathbf{u}) \ \ \ \ 
    \text{and}\ \  q(\mathbf{u})=-n_y(\mathbf{u})/n_z(\mathbf{u})
\]
We take three-pixel neighbourhoods $(\mathbf{u},\mathbf{v},\mathbf{w})$ to compute the gradient of $p,q$, where 
\[
    \frac{\partial p(\mathbf{u})}{\partial y}=p(\mathbf{w})-p(\mathbf{u})\ \text{,}\ 
    \frac{\partial q(\mathbf{u})}{\partial x}=q(\mathbf{v})-q(\mathbf{u})\ \ 
\]
In reality, due to noise and the discretisation to the pixel grid, the gradient field may not have exactly zero curl, but we seek the surface normals that give minimum curl values. Hence, the ternary cost is defined by:
\[
    \Psi(\mathbf{n}(\mathbf{u}),\mathbf{n}(\mathbf{v}),\mathbf{n}(\mathbf{w}))=\norm{p(\mathbf{w})-p(\mathbf{u})-(q(\mathbf{v})-q(\mathbf{u}))}.
\]

\paragraph{Graphical model optimisation}
We use higher order belief-propagation to minimise \eqref{eqn:GMcost} as implemented in the OpenGM toolbox\cite{andres2012opengm}. The optimum surface normal $\mathbf{n^\prime}$ will be labeled as one of the six possible disambiguations and we update our specular mask $L$ according to:
\[
    L(\mathbf{u})=
    \begin{cases}
        0 &\text{if}\ \mathbf{n}(\mathbf{u})\in \mathpzc{D} \\
        1 &\text{if}\ \mathbf{n}(\mathbf{u})\in \mathpzc{S} \\
    \end{cases}.
\]
The surface normals that result from this disambiguation process are still noisy (they use only local information) and may be subject to low frequency bias meaning that integrating them into a depth map does not yield good results. Hence, in Section \ref{sec:LPDfP} we solve globally for depth, using the stereo depth map as a guide to remove low frequency bias.

\section{Albedo estimation with gradient consistency}
We now use the surface normals estimated by the graphical model optimisation to compute an albedo map. In principal, the albedo can be computed from these normals and the unpolarised intensity simply by rearranging \eqref{eqn:lamblaw}. However, this purely local estimation is unstable and noise in the normals leads to artefacts in the estimated albedo map. We propose a simple but very effective regularisation to resolve this problem. We encourage the gradient of the estimated albedo map to be similar to the gradient of the unpolarised intensities at points where the intensity gradient is above a threshold and zero elsewhere. In other words, we encourage the albedo gradients to be sparse and hence the albedo piecewise uniform.

The estimated albedo minimises the following energy function
\begin{equation}
    E(\mathbf{u})=E_{\text{\it Lamb}}(\mathbf{u})+\lambda_I E_{\text{\it smooth}}(\mathbf{u}).
\end{equation}
The first term penalises the difference between rendered Lambertian intensity and estimated unpolarised intensity:
\begin{equation}
    E_{\text{\it Lamb}}(\mathbf{u})=\norm{a(\mathbf{u}) \mathbf{n}^\prime\cdot(\mathbf{u})\mathbf{s}-I_d(\mathbf{u})}^2_2
\end{equation}
where $I_d$ is diffuse dominant pixels from the estimated unpolarisation intensity, $\alpha$ represents a pixel-wise albedo map, $\mathbf{n^\prime}$ is the optimum surface normal map from the previous section and $\mathbf{s}$ the  light source. We can easily choose the diffuse pixels by excluding the specular mask where $L(\mathbf{u})=1$. 

\begin{figure}[t]
    \centering
    \includegraphics[width=8.5cm,clip=true]{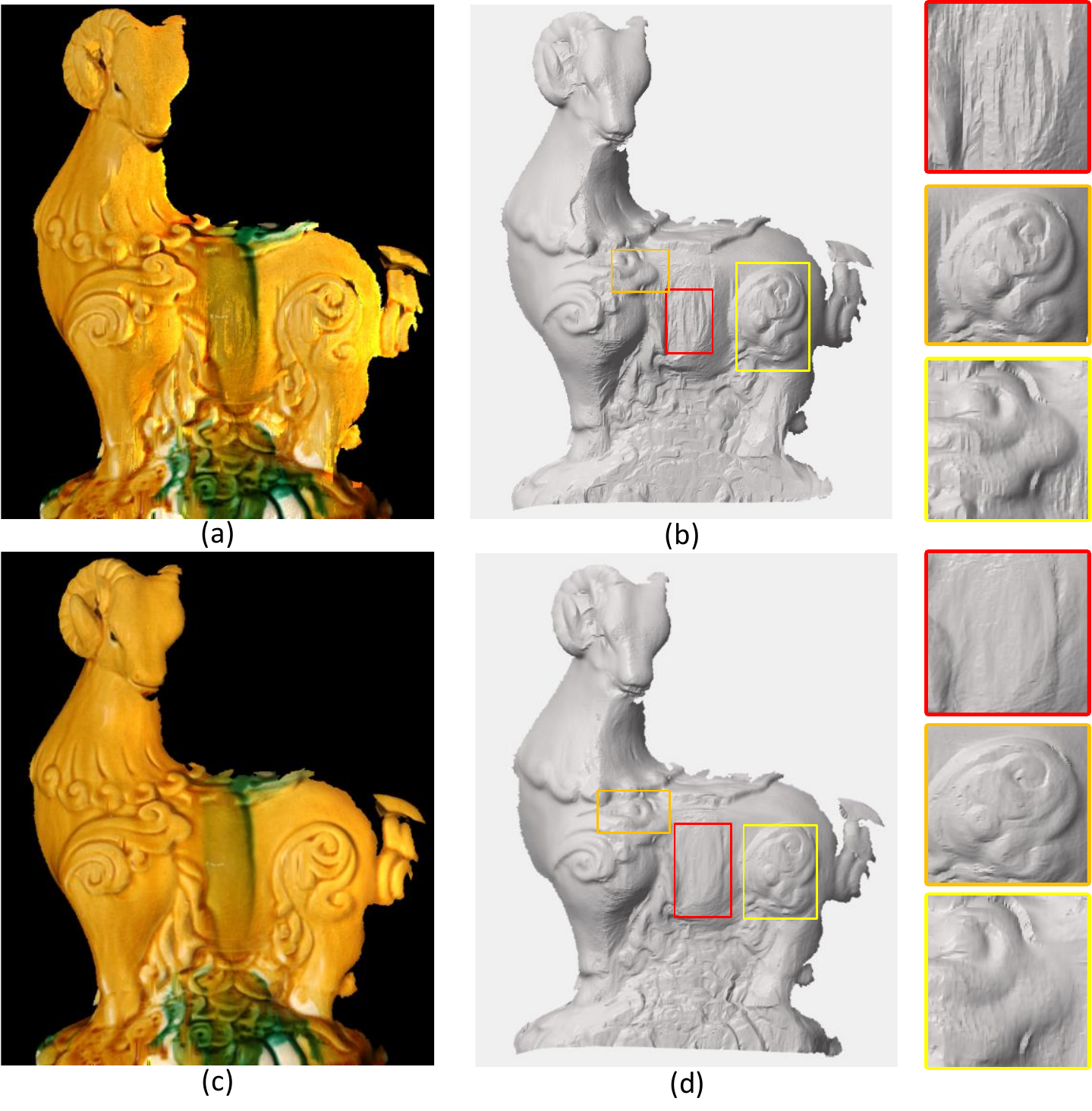}
    \caption{(a)/(c) Estimated albedo (b)/(d) Estimated geometry. First row: $\lambda_I=0$, second row: $\lambda_I=3$. Comparing (a) and (c), the albedo map becomes smoother. Comparing (b) and (d), the red rectangle region becomes smoother but while fine detailis largely preserved. }
    \label{fig:albedo_estimation}
\end{figure}

The second term penalises the difference between the estimated albedo gradient and the sparsified unpolarised intensity gradient. We denote the neighbour of $\mathbf{u}$ in $x$ direction with $v$ and $y$ direction with $w$, thus the smooth term can be written as
\begin{equation}
\begin{aligned}
         E_{\text{\it smooth}}(\mathbf{u})&=\norm{a(\mathbf{u})-a(\mathbf{v})-g(I_d(\mathbf{u})-I_d(\mathbf{v}))}\\
        &+\norm{a(\mathbf{u})-a(\mathbf{w})-g(I_d(\mathbf{u})-I_d(\mathbf{w}))}
    \end{aligned}
\end{equation}
where $g(.)$ is a threshold function that returns 0 if the input is $<t$, otherwise it returns the input albedo map only contains values on the diffuse pixels, we fill the hole on specular pixels with nearest neighbour method.
In Figure \ref{fig:albedo_estimation} we see how the smoothness term affects the estimated albedo map and depth. 

\section{Linear perspective depth from polarisation}\label{sec:LPDfP}

Finally, with albedo known and coarse depth values from two view stereo, we are ready to estimate dense depth from polarisation. We generalise a perspective camera model from Smith \etal \cite{smith2016linear}, note that it differs via the use of the coarse depth values and optimum normal from section \ref{sec:graph}. The fact that we estimate metric depth rather than relative height. As in \cite{smith2016linear}, we express polarisation and shading constraints in the form of a large, sparse linear system in the unknown depth values, meaning the method is very efficient and guaranteed to attain the globally optimal solution.

\paragraph{Phase angle constraint.}
The first constraint encourages the recovered surface normal to satisfy equation \eqref{eqn:phase}.
Following \cite{smith2016linear}, the projection of the surface normal into the image plane $(n_x,n_y)$ should be collinear with the phase angle vector. We seperate pixels into diffuse dominant and specular dominant with the help of specular mask $L$. The phase angle constraint for diffuse dominant pixels and specular dominant pixels are represented in first row and second row respectively in this matrix form:
\begin{equation}\label{eq:phaseconstraint}
    \begin{bmatrix}
         \cos(\phi(\mathbf{u})) & -\sin(\phi(\mathbf{u})) & 0\\
         \cos(\phi(\mathbf{u})+\frac{\pi}{2}) & -\sin(\phi(\mathbf{u})+\frac{\pi}{2}) & 0
    \end{bmatrix}
    \begin{bmatrix}
         n_x(\mathbf{u})\\ n_y(\mathbf{u})\\ n_z(\mathbf{u})
    \end{bmatrix}=0
\end{equation}

\paragraph{Shading/polarisation ratio constraint.}
Recall that the viewing angle is the angle between the surface normal and the viewer direction. Making the normalisation factor of the surface normal explicit, we can write $\cos(\theta_r(\mathbf{u}))=\frac{\mathbf{n}(\mathbf{u})\cdot \mathbf{v}(\mathbf{u})}{\norm{\mathbf{n}(\mathbf{u})}}$. By isolating the normalisation factor we arrive at:
\begin{equation}
    \norm{\mathbf{n}(\mathbf{u})}=\frac{\mathbf{n}(\mathbf{u})\cdot \mathbf{v}(\mathbf{u})}{\cos(\theta_r(\mathbf{u}))}.
\end{equation}
Substituting this into \eqref{eqn:lamblaw} we obtain:
\begin{equation} \label{eq:phase_constraint}
    \frac{\mathbf{n}(\mathbf{u})\cdot \mathbf{v}(\mathbf{u})}{\cos(\theta_r(\mathbf{u}))}=\frac{a(\mathbf{u}) \mathbf{n}(\mathbf{u})\cdot \mathbf{s}}{i_{\textrm{un}}(\mathbf{u})}
\end{equation}
Notice that our shading constraint only submit on the diffuse pixels. So we choose the pixels $\mathbf{u} \in \mathpzc{D}$ where $L(\mathbf{u})=0$. Unlike \cite{smith2016linear}, the perspective model means that the view vectors depend on pixel locations. Now we can reformulate the equation into a compact matrix form with respect to the surface normal:
\begin{equation} \label{eq:shadingvaries}
    \begin{bmatrix}
        s_x\cdot a(\mathbf{u})\cos{\theta(\mathbf{u})}-i_{\textrm{un}}(\mathbf{u})v_x(\mathbf{u})\\
        s_y\cdot a(\mathbf{u})\cos{\theta(\mathbf{u})}-i_{\textrm{un}}(\mathbf{u})v_y(\mathbf{u})\\
        s_z\cdot a(\mathbf{u})\cos{\theta(\mathbf{u})}-i_{\textrm{un}}(\mathbf{u})v_z(\mathbf{u})\\
    \end{bmatrix}^T
    \begin{bmatrix}
         n_x(\mathbf{u})\\ n_y(\mathbf{u})\\ n_z(\mathbf{u})
    \end{bmatrix}=0
\end{equation}

\paragraph{Surface normal constraint.}
We also encourage our recovered surface normal should co-linear with the optimised normal $n^\prime$ from section \ref{sec:graph} where their cross product is a zero vector. It can be formalised in following manner
\begin{equation} \label{eq:guide_n_constraint}
    \begin{bmatrix}
        0 &-n^\prime_z(\mathbf{u}) &n^\prime_y(\mathbf{u})\\
        n^\prime_z(\mathbf{u}) &0 &-n^\prime_x(\mathbf{u})\\
        -n^\prime_y(\mathbf{u}) &n^\prime_x(\mathbf{u}) &0
    \end{bmatrix}
    \begin{bmatrix}
         n_x(\mathbf{u})\\ n_y(\mathbf{u})\\ n_z(\mathbf{u})
    \end{bmatrix}=
    \begin{bmatrix}
        0\\0\\0
    \end{bmatrix}
\end{equation}

\paragraph{Global linear depth estimation.}
The relationship between the surface normal and depth under perspective viewing is given by \eqref{eq:perspectivenormal}. We can arrive at a linear relationship between the constraints described above and the unknown depth. 

We first extend \eqref{eq:perspectivenormal} to the whole image. Consider an image with $N$ foreground pixels whose unknown depth values are vectorised in $\mathbf{Z}\in\R^N$. The surface normal direction (unnormalised) can be computed for all pixels with:
\small
\begin{equation}
    \mathbf{NZ} = \begin{bmatrix}n_x(\mathbf{u}_1)\\\dots\\n_x(\mathbf{u}_N)\\n_y(\mathbf{u}_1)\\\dots\\n_y(\mathbf{u}_N)\\n_z(\mathbf{u}_1)\\\dots\\n_z(\mathbf{u}_N)\end{bmatrix},\quad
    \mathbf{N} = 
    \begin{bmatrix}
    -f_y\mathbf{I} & \mathbf{0} & \mathbf{0} \\
    \mathbf{0} & -f_x\mathbf{I} & \mathbf{0} \\
    \mathbf{X} & \mathbf{Y} & \mathbf{I} \\
    \end{bmatrix}
    \begin{bmatrix}\mathbf{D}_x \\ \mathbf{D}_y \\ \mathbf{I} \end{bmatrix}\label{eqn:perspnormim}
\end{equation}
\normalsize
where $\mathbf{X}=\textrm{diag}(x_1-x_0,\dots,x_N-x_0)$ and $\mathbf{Y}=\textrm{diag}(y_1-y_0,\dots,y_N-y_0)$. $\mathbf{D}_x, \mathbf{D}_y\in\R^{N\times N}$ compute finite difference approximations to the derivative of $Z$ in the $x$ and $y$ directions respectively. In practice, we use smoothed central difference approximations where possible, reverting to central or forward/backward differences where all neighbours are not available. Hence $\mathbf{D}_x, \mathbf{D}_y$ have at most six non-zero values per row.

Combining \eqref{eqn:perspnormim} with \eqref{eq:phaseconstraint}, \eqref{eq:shadingvaries} and \eqref{eq:guide_n_constraint} leads to equations that are linear in depth.
We now combine these equations into a large linear system of equations for the whole image.
Of the $N$ foreground pixels we divide these into diffuse and specular pixels according to the mask $L$. We denote the number of diffuse pixels with $N_D$ and specular with $N_S$. We now form a linear system in the vector of unknown depth values, $\mathbf{Z}$:
\begin{equation}\label{eqn:linsystem}
    \begin{bmatrix}
    \lambda\mathbf{AN} \\
    \mathbf{W}
    \end{bmatrix}\mathbf{Z} = \begin{bmatrix}
    \mathbf{0}_{4N+N_D} \\
    Z_{\textrm{guide}}(\mathbf{u}_1) \\
    \vdots \\
    Z_{\textrm{guide}}(\mathbf{u}_N)
    \end{bmatrix}
\end{equation}
where $Z_{\textrm{guide}}(\mathbf{u}_i)$ are the stereo depth values from Section \ref{sec:graph} and $\mathbf{W}\in\R^{K\times N}$ performs a sparse indices matrix of ${\bm Z}$ at positions $(x_1,y_1),\dots,(x_K,y_K)$. $\mathbf{I}_N\in\R^{N\times N}$ is the identity matrix and $\mathbf{0}_{4N+N_D}$ is the zero vector of length $4N+N_D$. $\mathbf{A}$ has $4N+N_D$ rows, $3N$ columns and is sparse. Each row evaluates one equation of the form of \eqref{eq:phaseconstraint}, \eqref{eq:shadingvaries} and \eqref{eq:guide_n_constraint}. $\lambda>0$ is a weight which trades off the influence of the guide depth values against satisfaction of the polarisation constraints. We then solve \eqref{eqn:linsystem} in a least squares sense using sparse linear least squares.

\section{Experimental results}

\begin{figure}[t] 
    \centering
    \includegraphics[width=8.5cm,clip=true]{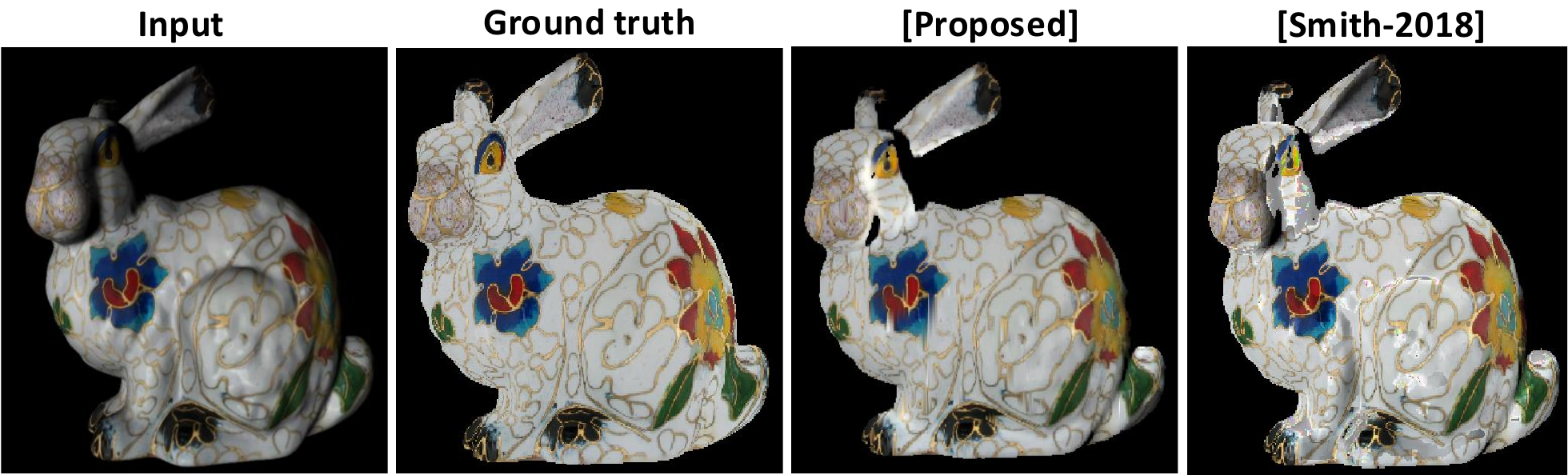}
    \caption{Albedo estimates on synthetic data.}
    \label{fig:synthetic_comparison_albedo}
\end{figure}

We present experimental results on both synthetic and real data. We compare our method against \cite{hirschmuller2005accurate,smith2016linear,smith2018height,kadambi2015polarized,wu2014real}, the differences are summarised in Table \ref{tab:methods}. We set $\lambda_I=1, \lambda=1$ and $t=0.01$ through our experiments. Note that the source code for \cite{kadambi2015polarized} is not available so we are only able to compare against a single result provided by the authors. Similarly, real image results for \cite{wu2014real} were provided by the author running the implementation for us. Whereas \cite{hirschmuller2005accurate,smith2016linear,smith2018height} are open sourced and we compare quantitatively.
For synthetic data, we render images of the Stanford bunny with Blinn-Phong reflectance with varying albedo texture using the pinhole camera model, as shown in Figure \ref{fig:synthetic_comparison_albedo} (left). The texture map is from \cite{Texturemontage05}. We simulate the effect of polarisation according to \eqref{eqn:TRS} by setting refractive index value to $1.4$ and corrupt the polarisation image and second camera intensity by adding Gaussian noise with zero mean and standard deviation $\sigma$. The metric ground truth of the depth map is range between $72.33$mm to $90.09$mm.

\begin{table}[!t]
    \centering
\resizebox{\columnwidth}{!}{%
    \begin{tabular}{|c||c|c|c|}
    \hline
         &Coarse depth & Shading & Polarisation \\
                \hline
         Stereo\cite{hirschmuller2005accurate} & \checkmark & & \\ \hline
         Smith-2016\cite{smith2016linear} & & \checkmark &\checkmark \\ \hline
         Smith-2018\cite{smith2018height} & & \checkmark &\checkmark \\ \hline
         Polarised 3D \cite{kadambi2015polarized} & \checkmark & & \checkmark \\ \hline
         Wu-2014 \cite{wu2014real} & \checkmark & \checkmark &  \\ \hline
         Proposed & \checkmark & \checkmark & \checkmark\\ \hline
    \end{tabular}
    }%
    \caption{Summary of the different method}
    \label{tab:methods}
\end{table}
\begin{table}[h]\centering
    {\footnotesize
    \setlength{\tabcolsep}{2pt}
        \begin{tabular}{|c|c|c|c|c|c|c|c|c|}
        \hline
         & \multicolumn{2}{c|}{$\sigma=0\%$} & \multicolumn{2}{c|}{$\sigma=0.5\%$} & \multicolumn{2}{c|}{$\sigma=1\%$} \\ \cline{2-7}
         \multirow{2}{*}{\bf Method} &Depth &Normal &Depth &Normal &Depth &Normal\\ 
         & (mm) &(deg) &(mm) &(deg) &(mm) &(deg)\\
        \hline
        \hline
        \cite{hirschmuller2005accurate} &0.49 &38.151 &0.49 &39.78 &0.49 &39.67\\
        \cline{1-7}
        \cite{smith2016linear} &10.68 &30.38 &85.91 &29.966 &113.80 &32.03\\ \cline{1-7}
        \cite{smith2018height} &12.02 &22.53 &36.08 &26.54 &40.88 &28.54\\
        \cline{1-7}
         Prop & 0.29 & 9.799 &0.30 &9.86 &0.31 & 14.03\\ \cline{1-7}
        \end{tabular}
    }
    \caption{Mean absolute difference in depth and mean angular surface normal errors on synthetic data. For \cite{smith2016linear,smith2018height} methods reconstructed the depth up to scale we compute the optimum scale to align with the ground truth depth map.}
    \label{tab:quan}
\end{table}

\begin{figure}[!t] 
    \centering
    \includegraphics[width=8.5cm,clip=true]{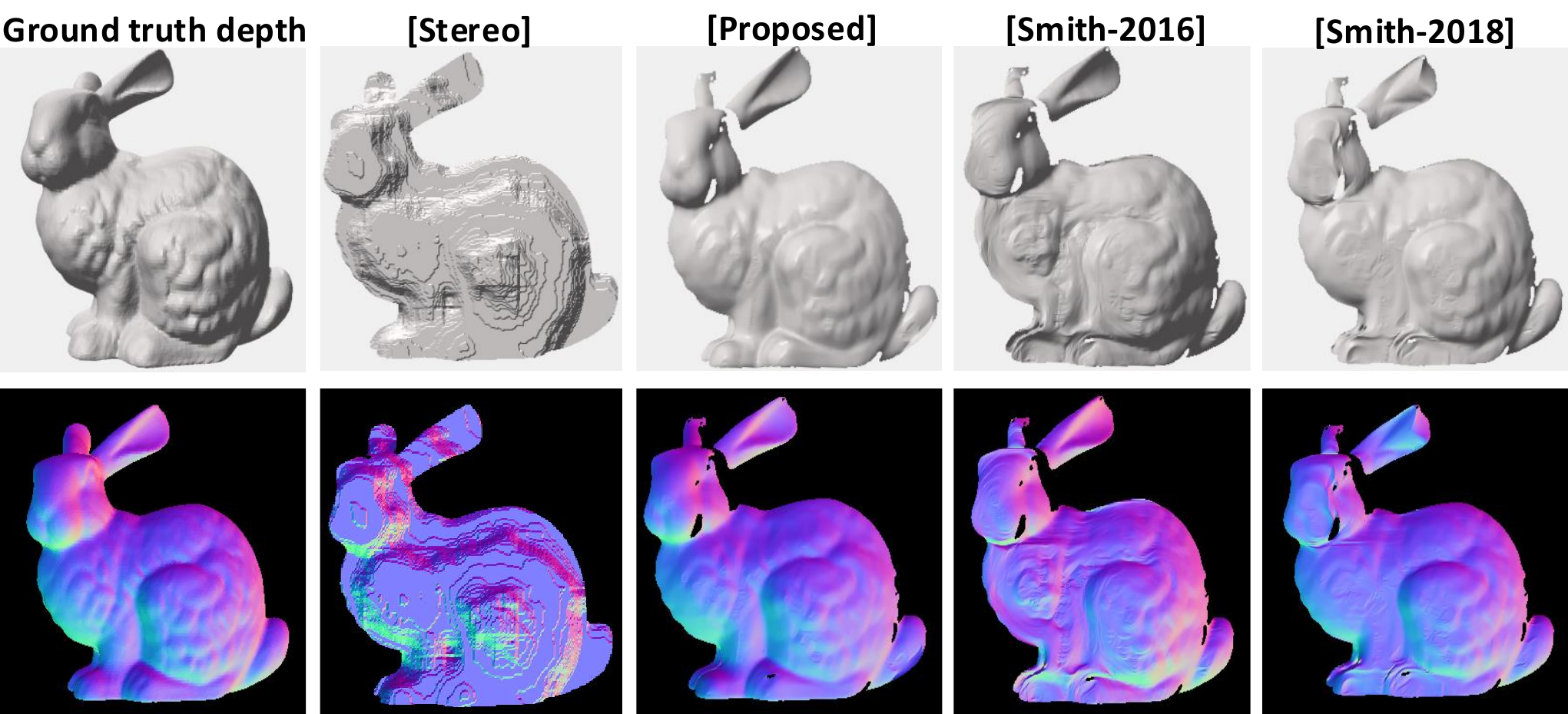}
    \caption{Qualitative shape estimation results on synthetic data with comparison with \cite{smith2016linear}}
    \label{fig:synthetic_comparison_depthnormal}
\end{figure}

In Figure \ref{fig:synthetic_comparison_albedo} we show the estimated albedo map of the synthetic data and compare with \cite{smith2018height}. In Table \ref{tab:quan} we show the mean absolute error in the surface depth (in millimetre) and mean angular error (in degrees) in the surface normals. We include comparison with the initial stereo depth \cite{hirschmuller2005accurate} and state-of-the-art polarisation methods \cite{smith2016linear,smith2018height}. In Figure \ref{fig:synthetic_comparison_depthnormal} we display the qualitative results of this experiment. 

\begin{figure*}[!t]
    \centering
    \includegraphics[width=18cm,clip=true]{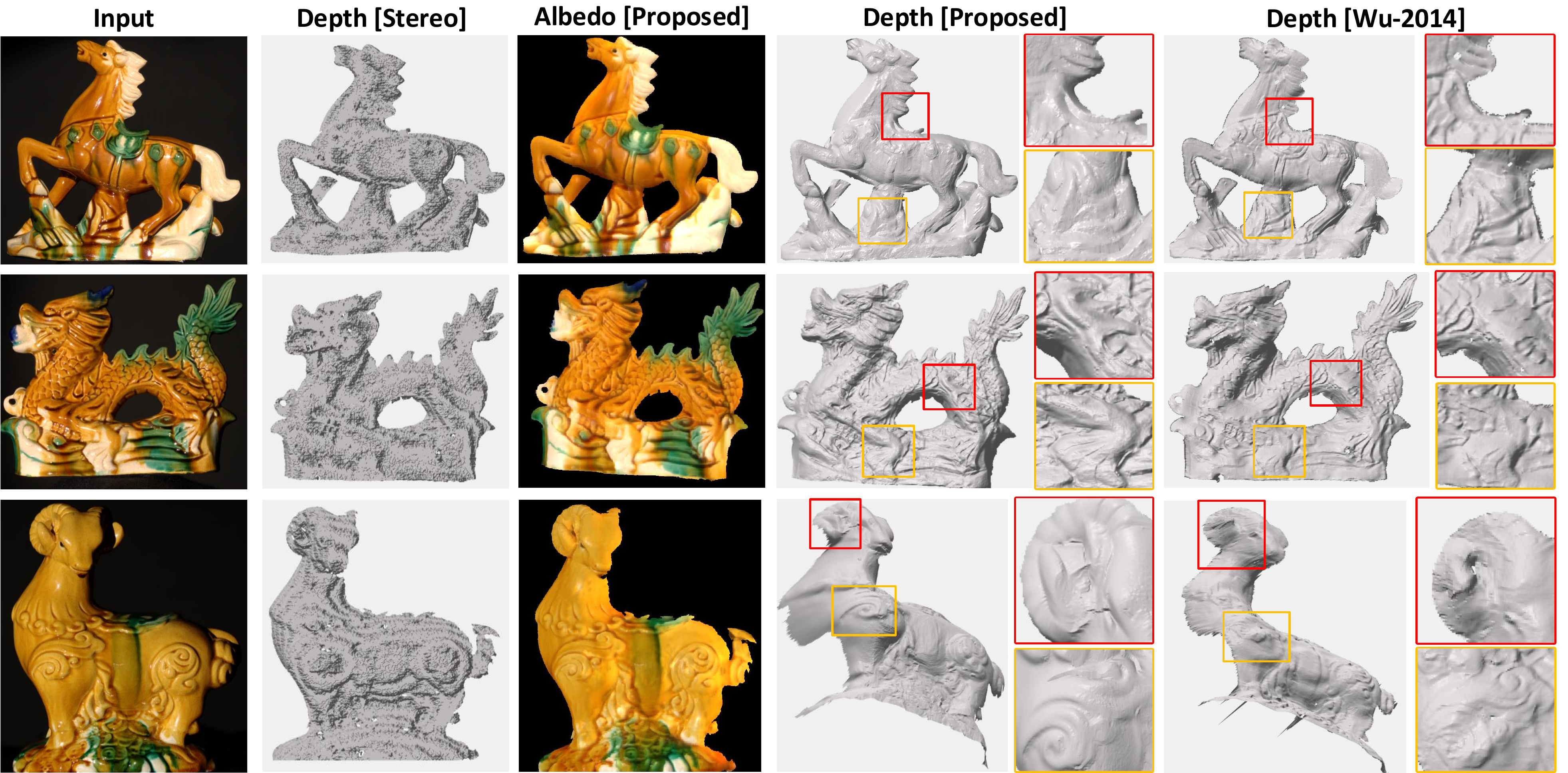}
    \caption{We show our results on complex object. From left to right we show an image from the input sequence; Depth from stereo reconstruction\cite{hirschmuller2005accurate}; Our proposed estimated albedo map and the estimated depth. Depth estimation by \cite{wu2014real}.}
    \label{fig:Results}
\end{figure*}

Next we show results on a dataset of real images. The first dataset is from \cite{kadambi2015polarized}. Although the depth here is provided by a Kinect sensor, not stereo, our graphical model optimisation in Section \ref{sec:graph} can take any source of depth map. In this case we replace the depth map with the Kinect one and keep the rest of the process identical when we evaluate the data. The comparison can be viewed in Figure \ref{fig:Polarised3D_comparison} where we show that our proposed result can give more details on the reconstruction. In this experiment, we estimate the light source direction using \cite{smith2016linear}.

We then show results on our own collected data. We place the polarisation and RGB cameras with parallel image planes and the RGB camera shifted 5cm along the $x$ axis relative to the polarisation camera as illustrated in Figure \ref{fig:teaser}. We compare our method with \cite{wu2014real} directly performed by the author. In Figure \ref{fig:Results} we show qualitative results for three objects with glossy reflectance and varying albedo. Our method gives improved detail (see insets) but also more stable overall shape (see third row). Notice that in this experiment we calibrated the light source in advance with a uniform albedo sphere using method in \cite{smith2016linear}.

\section{Conclusions}

In this paper we have proposed a method for estimating dense depth and albedo maps for glossy, dielectric objects with varying albedo. We do so using a hybrid imaging system in which a polarisation image is augmented by a second view from a standard RGB camera. We avoid assumptions common to recent methods (constant albedo, orthographic projection) and reduce low frequency distortion in the recovered depth maps through the stereo cue.

\begin{figure}[t] 
    \centering
    \includegraphics[width=8.5cm,clip=true]{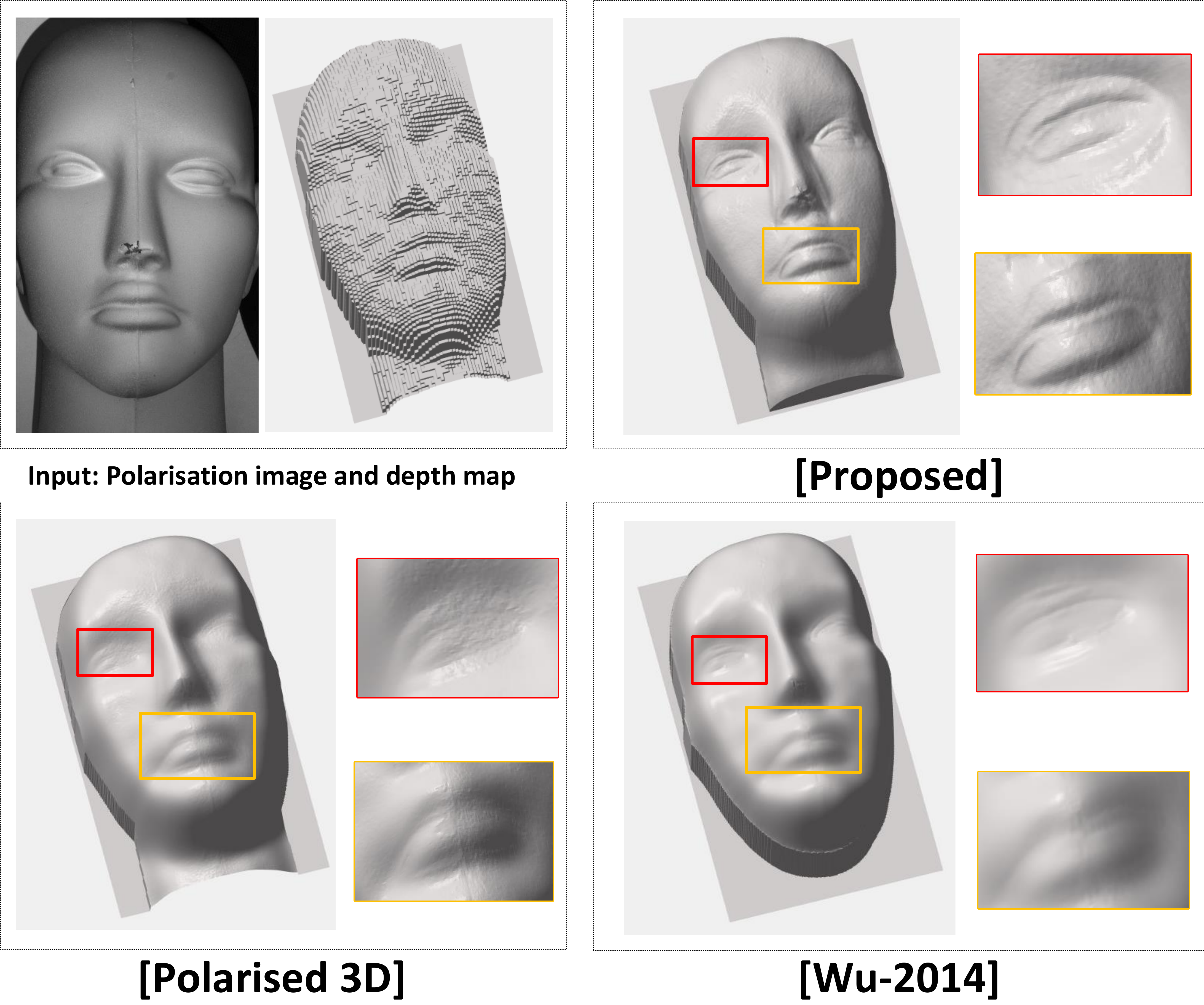}
    \caption{Comparison on \cite{kadambi2015polarized} dataset. Top-left: One of the polarisation intensity images and Kinect depth map. Top-right: our result. Bottom-Left: \cite{kadambi2015polarized}. Bottom-Right:  \cite{wu2014real}.}
    \label{fig:Polarised3D_comparison}
\end{figure}

Since we rely on stereo, our method does not work well on  textureless objects. However, note that our method works equally well with a Kinect depth map as the result shows in Figure \ref{fig:Polarised3D_comparison}. We also assume the refractive index is known in our framework. It could be potentially measured given a sufficiently accurate guide depth map. Although our stereo setup cannot provide this, it could potentially be provided by photometric stereo or multiview stereo.
There are many exciting possibilities for extending this work. The lighting, reflectance and polarisation models could be generalised. In particular, a more comprehensive model of mixed specular/diffuse reflectance and polarisation would be beneficial. Our linear approach is efficient and does not require initialisation, but it may be useful to subsequently perform a nonlinear optimisation over all unknowns (depth, albedo, refractive index) simultaneously such that the true underlying objective function can be minimised (taking inspiration from \cite{yu2017shape}).

\clearpage
{\small
\bibliographystyle{ieee}
\bibliography{egbib}
}

\end{document}